 \documentclass[final,3p,times,twocolumn]{elsarticle}

\usepackage{amssymb}
\usepackage{amsmath}
\usepackage{breqn}

\usepackage{lineno}
\usepackage{natbib}
\usepackage{subcaption}
\usepackage{multirow}
\usepackage{balance}


\begin{document}

\begin{frontmatter}



\title{Recurrent neural networks with specialized word embeddings \\ for health-domain named-entity recognition}


\author[UTS,CMCRC]{Inigo Jauregi Unanue}
\author[CMCRC]{Ehsan Zare Borzeshi}
\author[UTS]{Massimo Piccardi}

\address[UTS]{University of Technology Sydney (UTS), Sydney, Australia}
\address[CMCRC]{Capital Markets Cooperative Research Center (CMCRC), Sydney, Australia}

\begin{abstract}
\textbf{Background.} Previous state-of-the-art systems on Drug Name Recognition (DNR) and Clinical Concept Extraction (CCE) have focused on a combination of text ``feature engineering" and conventional machine learning algorithms such as conditional random fields and support vector machines. However, developing good features is inherently heavily time-consuming. Conversely, more modern machine learning approaches such as recurrent neural networks (RNNs) have proved capable of automatically learning effective features from either random assignments or automated word ``embeddings''.\\
\textbf{Objectives.} (i) To create a highly accurate DNR and CCE system that avoids conventional, time-consuming feature engineering. (ii) To create richer, more specialized word embeddings by using health domain datasets such as MIMIC-III. (iii) To evaluate our systems over three contemporary datasets.\\
\textbf{Methods.} Two deep learning methods, namely the Bidirectional LSTM and the Bidirectional LSTM-CRF, are evaluated. A CRF model is set as the baseline to compare the deep learning systems to a traditional machine learning approach. The same features are used for all the models.\\
\textbf{Results.} We have obtained the best results with the Bidirectional LSTM-CRF model, which has outperformed all previously proposed systems. The specialized embeddings have helped to cover unusual words in \textit{DrugBank} and \textit{MedLine}, but not in the \textit{i2b2/VA} dataset.\\
\textbf{Conclusion.} We present a state-of-the-art system for DNR and CCE. Automated word embeddings has allowed us to avoid costly feature engineering and achieve higher accuracy. Nevertheless, the embeddings need to be retrained over datasets that are adequate for the domain, in order to adequately cover the domain-specific vocabulary.
\end{abstract}


\end{frontmatter}


\section{Introduction}
\label{S:1}

In recent years, the amount of digital information generated from all sectors of society has increased rapidly, and as a result, agriculture, industry, small businesses and, of course, healthcare, are becoming more efficient and productive thanks to the insights obtained from the ``Big Data". However, in order to deal effectively with such large data, there is an ongoing need for novel, scalable and more accurate analytic tools.

\begin{figure*}[t!]
	\centering\includegraphics[width=\textwidth]{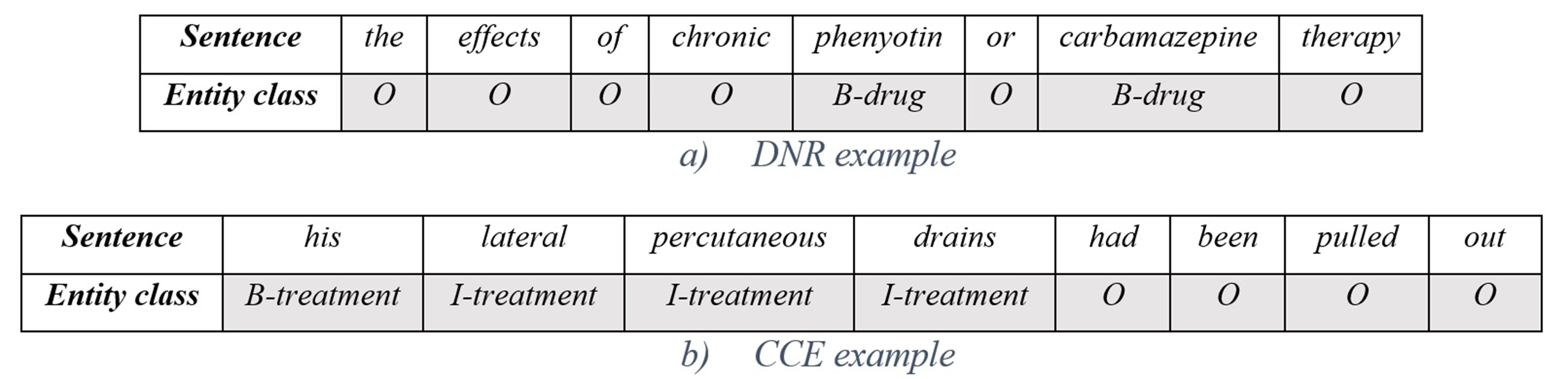}
	\caption{(a) DNR and (b) CCE tasks examples, where `B' (beginning) specifies the start of a named entity, `I' (inside) specifies that the word is part of the same named entity, and `O' (outside) specifies that the word is not part of any predefined class}
	\label{fig:fig1}
\end{figure*}

In the healthcare system, patients' medical records represent a big data source. Even though the records contain very useful information about the patients, in most cases the information consists of unstructured text such as, among others, doctors' notes, medical observations made by various physicians, and descriptions of the recommended treatments. This type of data cannot be analyzed using common statistical tools; rather, they need to be approached by Natural Language Processing (NLP) techniques. In this paper, we focus on a well-known task in NLP, namely Named-Entity Recognition (NER). The goal of NER is to automatically find ``named entities" in text and classify them into predefined categories such as people, locations, companies, time expressions etc. In the case of specialized domains, NER systems focus on text with specific dictionaries and topics, together with dedicated sets of named-entities. In the health domain, the two most important NER tasks are Clinical Concept Extraction (CCE) and Drug Name Recognition (DNR). The former aims to identify mentions of clinical concepts in patients' records to help improve the organization and management of healthcare services.  Named entities in CCE can include test names, treatments, problems related to individual patients, and so forth. The latter seeks to find drug mentions in unstructured biomedical texts to match drug names with their effects and discover drug-drug interactions (DDIs). DNR is a key step of pharmacovigilance (PV) which is concerned with the detection and understanding of adverse effects of drugs and other drug-related problems. Figure \ref{fig:fig1} shows examples of both tasks.

NER is a challenging learning problem because in most domains the training datasets are scarce, preventing a ``brute-force" approach by exhaustive dictionaries. Consequently, many systems rely on hand-crafted rules and language-specific knowledge to solve this task. To give a simple example of such rules, if the word begins with a capital letter in the middle of the sentence, it can be assumed to be a named entity in most cases. Nevertheless, these approaches are time-costly to develop, depend considerably on the language and the domain, are ineffective in the presence of informal sentences and abbreviations and, although they usually achieve high precision, suffer from low recall (i.e., they miss many entities). Conversely, machine learning (ML) approaches overcome all these limitations as they are intrinsically robust to variations. Current state-of-the-art ML methods follow a two-step process: 1) feature engineering and 2) automated classification \cite{segura2015exploring,abacha2015text,rocktaschel2013wbi,de2011machine}. The first step represents the text by numeric vectors using domain-specific knowledge. The second step refers to the task of classifying each word into a different named-entity class, with popular choices for the classifier being the linear-chain Conditional Random Fields (CRF), Structural Support Vector Machines (S-SVM) and maximum-entropy classifiers. The drawback of this approach is that feature engineering can be often as time-consuming as the manual design of rules.
 
In recent years, the advent of deep learning has contributed to significantly overcome this problem \cite{lample2016neural,hinton2012deep,krizhevsky2012imagenet}. The Long Short-Term Memory (LSTM) and its variants (e.g., the Bidirectional LSTM), which are a specific type of Recurrent Neural Networks (RNNs), have reported very promising results \cite{lample2016neural}. In these models, words only need to be assigned to random vectors, and during training the neural network is able to automatically learn improved representations for them, completely bypassing feature engineering. In order to further increase the performance of these systems, the input vectors can alternatively be assigned with general-purpose word embeddings learned with GloVe or Word2vec \cite{pennington2014glove,mikolov2013distributed}. The aim of general-purpose word embeddings is to map every word in a dictionary to a numerical vector (the embedding) so that the distance between the vectors somehow reflects the semantic difference between the words. For example, `cat' and `dog' should be closer in the vector space than `cat' and `car'. The common principle behind embedding approaches is that the meaning of a word is conveyed by the words it is used with (its surrounding words, or context). Therefore, the training of the word embeddings only requires large, general-purpose text corpora such as Wikipedia (400K unique words) or Common Crawl (2.2M unique words), without the need for any manual annotation. However, drug and clinical concept recognition are very domain-specific tasks, and many words might not appear in general-domain datasets. In order to assign word embeddings to these specialized words, the embedding algorithms need to be retrained using medical domain resources such as the MIMIC-III corpora \cite{johnson2016mimic}. As well as semantic word embeddings, \textit{character-level} embeddings of words can also be automatically learned. Such embeddings can capture typical prefixes and suffixes, providing the classifiers with richer representations of the words \cite{lample2016neural}.
 
Preliminary results for the work presented in this paper have obtained very promising accuracy in DNR and CCE tasks using neural networks. Chalapathy et al. \cite{chalapathy2016investigation} presented a DNR system that uses a Bidirectional LSTM-CRF architecture with random assignments of the input word vectors at the EMNLP 2016 Health Text Mining and Information Analysis workshop. The reported results were very close to the system that ranked first in the SemEval-2013 Task 9.1. In Chalapathy et al. \cite{chalapathy2016bidirectional}, the authors leveraged the same architecture for CCE at the Clinical NLP 2016 workshop, this time using pre-trained word embeddings from GloVe, and the results outperformed previous systems over the 2010 i2b2/VA IRB Review dataset. In this paper, we extend the previous research by training the deep networks with more complex and specialized word embeddings. Moreover, we explore the impact of augmenting the word embeddings with conventional feature engineering. As methods, we compare contemporary recurrent neural networks such as the Bidirectional LSTM and the Bidirectional LSTM-CRF against a conventional ML baseline (a CRF). We report state-of-the-art results in both DNR and CCE.

\section{Related work}
\label{S:2}

Most of the research carried out in domain-specific NER has combined supervised and semi-supervised ML models with text feature engineering. For example, the WBI-NER system that ranked first in the SemEval-2013 Task 9.1 (Recognition and classification of pharmacological substances, DNR) \cite{rocktaschel2013wbi}, is based on a linear-chain CRF with specialized features. Other similar systems for DNR \cite{abacha2015text,liu2015feature} use various general- and domain-specific features. In CCE, the same approach (feature engineering + conventional ML classifier) has achieved the best results \cite{de2011machine,boag2015cliner}.

In the recent years, there has been an increase in the use of deep neural networks for a variety of NLP tasks, including NER \cite{lample2016neural,hinton2012deep,krizhevsky2012imagenet}. Pre-trained word embeddings \cite{pennington2014glove,mikolov2013distributed,lebret2013word} have been used in traditional ML methods \cite{nikfarjam2015pharmacovigilance,wu2015study} and in neural networks, where Deconourt et al. \cite{dernoncourt2017identification} has achieved better performance than previously published systems in de-identification of patient notes. Cocos et al. \cite{cocos2017deep} have used the Bidirectional LSTM model for labelling Adverse Drug Reactions in pharmacovigilance. Xie et al. \cite{xie2017mining} have used a similar model for studying the adverse effects of e-cigarettes. Wei et al. \cite{wei2016disease} have combined the output of a Bidirectional LSTM and a CRF as input to an SVM classifier for disease name recognition. A possible drawback of this approach is that the overall prediction is not structured and may miss on useful correlation between the output variables.

In a work that is more related to ours, Jaganatha and Yu \cite{jagannatha2016structured} have employed a Bidirectional LSTM-CRF to label named entities from electronic health records of cancer patients. Their model differs in the CRF output module where the pairwise potentials are modelled using a Convolutional Neural Network (CNN) rather than the usual transition matrix. Gridach \cite{gridach2017character} has also used the Bidirectional LSTM-CRF for named-entity recognition in the biomedical domain.

The main difference and contribution of the proposed approach is that it leverages specialized health-domain embeddings created from a structured database. In the experiments, these embeddings have been used jointly with general-domain embeddings and they have proved able to improve the accuracy in several cases. In addition, our work evaluates the use of hand-crafted features in the system \cite{lee2016feature}. This aims to provide a comprehensive feature comparison for health-domain named-entity recognition based on LSTM models.

\section{Methods}
\label{S:3}

In this section we provide a description of the main methods employed. First, we describe the conditional random field (CRF), a traditional machine learning approach for the classification of sequences, which is used as a baseline in the experiments. This baseline is compared with two variants of a contemporary recurrent neural network, which are known as Bidirectional LSTM and Bidirectional LSTM-CRF, respectively.

\begin{figure*}[t!]
	\centering\includegraphics[width=\textwidth]{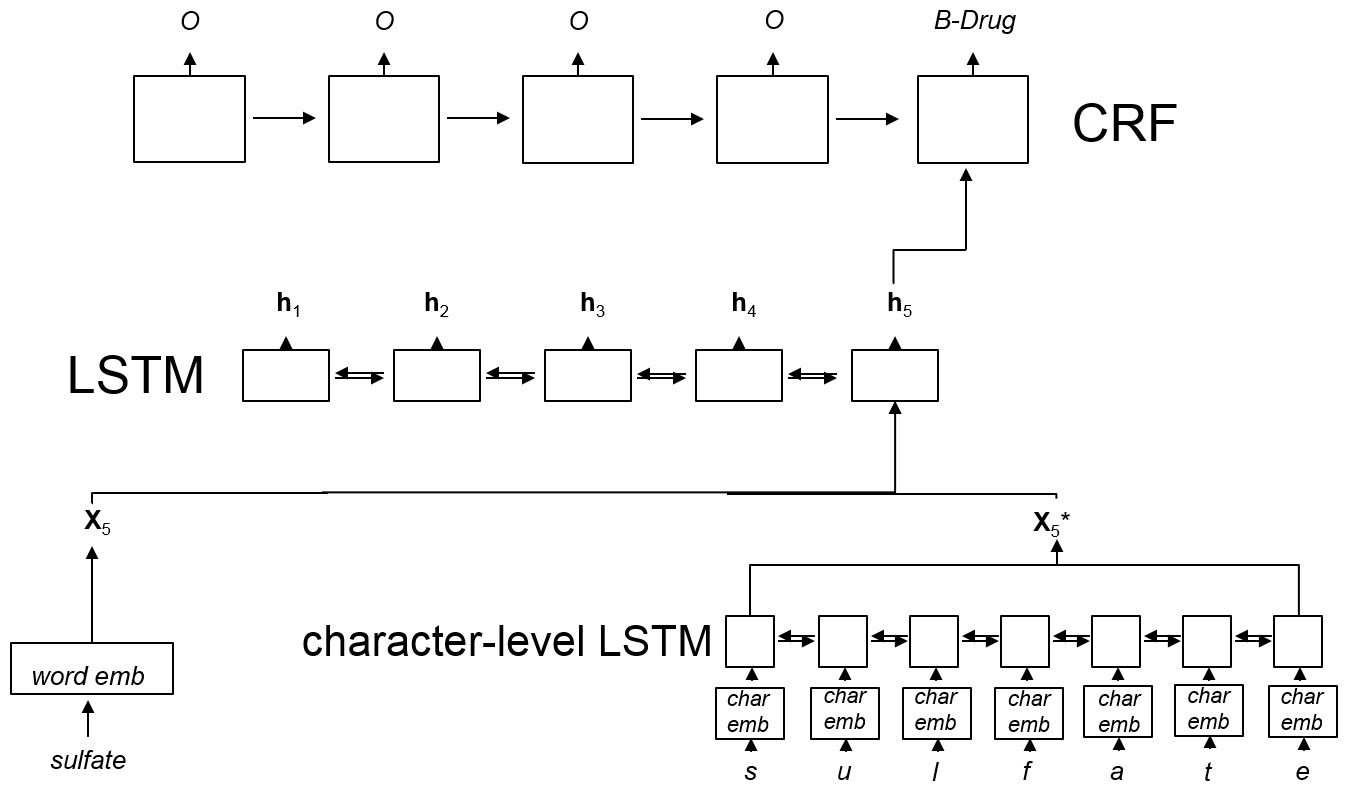}
	\caption{The Bidirectional LSTM-CRF with word-level and character-level word embeddings. In the example, word `sulfate' is assumed to be the 5th word in a sentence and its only entity; `\textbf{x}_5' represents its word-level embedding (a single embedding for the whole word); `{\textbf{x}_5}^*' represents its character-level embedding, formed from the concatenation of the last hidden state of the forward and backward passes of a character-level Bidirectional LSTM; 'h_{1–5}' are the hidden states of the main Bidirectional LSTM which become the inputs into a final CRF; eventually, the CRF provides the labeling.}
	\label{fig:fig_diagram}
\end{figure*}

\subsection{CRF}

A CRF model is a well-known ML approach that has been widely used in NER \cite{lafferty2001conditional}. It predicts sequences of labels ($\textbf{y}$) from sequences of measurements ($\textbf{x}$) taking into account the sequentiality of the data. In a CRF model, $p(\textbf{y}|\textbf{x},\textbf{w})$ is given in Eq. \ref{eq:CRF_1} below, where $\textbf{w}$ notes the model's parameters, $\psi(\textbf{x},\textbf{y})$ is the chosen feature vector and $Z(\textbf{w},\textbf{x})$ is the cumulative sum of $p(\textbf{y}|\textbf{x},\textbf{w})$ over all the possible $\textbf{y}$:

\begin{equation}
\label{eq:CRF_1}
p(\textbf{y}|\textbf{x},\textbf{w})=\frac{exp(\textbf{w}^T \psi(\textbf{x},\textbf{y}))}{Z(\textbf{w},\textbf{x})}
\end{equation}

The parameters of this model are typically learned from a training set, $(\textbf{\textbf{Y},\textbf{X}})=\{\textbf{x}_i,\textbf{y}_i\}$, $i= 1 \dots N$, with conditional maximum likelihood as in:

\begin{equation}
\label{eq:CRF_parameters}
\textbf{w}=\arg\max_{\textbf{w}}p(\textbf{Y}|\textbf{X},\textbf{w})
\end{equation}

Once the model has been trained, the prediction of a CRF is the sequence of labels maximizing the model for the given input sequence and the learned parameters:

\begin{equation}
\label{eq:CRF_inference}
\textbf{y}^*=\arg\max_{\textbf{y}}p(\textbf{y}|\textbf{x},\textbf{w})
\end{equation}

The labels are typically predicted using a Viterbi-style algorithm which provides the optimal prediction for the measurement sequence as a whole. The model is trained by maximizing the conditional likelihood, or cross-entropy, over a given training set. For its implementation, we have used the HCRF library \cite{HCRF}. The features used as input are described in Section \ref{S:4}. In the experiments, we use the CRF as a useful baseline for performance comparison with the proposed neural networks. Note that a CRF model is also used as the output layer in the Bidirectional LSTM-CRF as explained in the next section.

\subsection{Bidirectional LSTM and bidirectional LSTM-CRF}

RNNs are a type of neural network architecture in which connections between units form a directed cycle, creating an internal state and achieving dynamic temporal behavior. Thanks to their internal memory, RNNs can process a sequence of vectors $(\textbf{x}_1, \textbf{x}_2,..., \textbf{x}_n)$ as input and produce another sequence $(\textbf{h}_1, \textbf{h}_2,..., \textbf{h}_n)$ as output that contains some extent of sequential information about every vector in the input. However, these architectures in practice fail to learn long-term dependencies in the sequences as they tend to be biased by the most recent vectors \cite{bengio1994learning}. The Long Short-Term Memory (LSTM) was therefore designed to overcome this issue by incorporating a gated memory-cell that has been shown to capture long-term dependencies \cite{hochreiter1997long}. Eq. \ref{eq:LSTM} shows the implementations of the different gates in the LSTM \cite{lample2016neural}, where $\textbf{i}_t$ is the ``input" gate, $\textbf{c}_t$ is the ``cell" gate, $\textbf{o}_t$ is the ``output" gate, $\textbf{W}$ are the weights of the network, $\textbf{b}$ are the biases, $\sigma$ is the element-wise sigmoid function, and $\odot$ is the element-wise product. The bidirectional LSTM (B-LSTM) is just a variation, in which both the left-to-right $(\overrightarrow{\textbf{h}_{t}})$ and the right-to-left $(\overleftarrow{\textbf{h}_{t}})$ representations of the input sentence are generated, and then concatenated $\textbf{h}_{t}=[\overrightarrow{\textbf{h}_{t}};\overleftarrow{\textbf{h}_{t}}]$ in order to obtain the final representation.

\begin{equation}
\label{eq:LSTM}
\begin{split}
\textbf{i}_{t}  &= \sigma(\textbf{W}_{xi}\textbf{x}_{t}+\textbf{W}_{hi}\textbf{h}_{t-1}+\textbf{W}_{ci}\textbf{c}_{t-1}+\textbf{b}_{i})\\
\textbf{c}_{t}  &= (1-\textbf{i}_{t})\odot \textbf{c}_{t-1}+\\
&\textbf{i}_{t} \odot \tanh(\textbf{W}_{xc}\textbf{x}_{t}+\textbf{W}_{hc}\textbf{h}_{t-1}+\textbf{b}_{c})\\
\textbf{o}_{t}  &= \sigma(\textbf{W}_{xo}\textbf{x}_{t}+\textbf{W}_{ho}\textbf{h}_{t-1}+\textbf{W}_{co}\textbf{c}_{t}+\textbf{b}_{o})\\
\textbf{h}_{t}  &= \textbf{o}_{t}\odot \tanh(\textbf{c}_{t}) 
\end{split}
\end{equation}

When applying the LSTM in NER, the words in the input sentence are first mapped to numerical vectors. These vectors can be random valued, a pre-trained word embedding, domain-specific word features or any combination of them. For each vector, the output of the network are the posterior probabilities of each named-entity class. An improvement of these networks has been presented by Lample et al. \cite{lample2016neural} using a CRF as a final output layer. This final layer provides the system with the ability to perform joint decoding of the input sequence in a Viterbi-style manner. The resulting network is known as the Bidirectional LSTM-CRF (B-LSTM-CRF). We test the LSTM models with the same features used for the CRF in order to establish the fairest-possible comparison. The features are described in detail in Section \ref{S:4}. Fig.\ref{fig:fig_diagram} shows a descriptive diagram of the B-LSTM-CRF.

\section{Word features}
\label{S:4}

As mentioned above, neural networks can learn meaningful representations from random inizializations of word embeddings. However, it has been proved that pre-trained word embeddings can improve the performance of the network \cite{lample2016neural,chalapathy2016bidirectional,dernoncourt2017identification,lee2016feature}. In this section, we present the pre-trained embeddings employed in lieu of the random assignments. 

\subsection{Specialized word embeddings}

A word embedding maps a word to a numerical vector in a vector space, where semantically-similar words are expected to be assigned similar vectors. To perform this mapping, we have used a well-known algorithm called GloVe \cite{pennington2014glove}. This algorithm learns word embeddings by looking at the co-occurrences of the word in the training data, assuming that a word's meaning is mostly defined by its context and, therefore, words having similar contexts should have similar embeddings. GloVe can be trained from large, general-purpose datasets such as Wikipedia, Gigaword5 or Common Crawl without the need for any manual supervision. In this work, we have experimented with different general-purpose, pre-trained word embeddings from the official GloVe website \cite{GloVe} and noticed that the embeddings trained with Common Crawl (cc) (2.2M unique words) were giving the best results. We have employed these embeddings on their own, and also concatenated with the MIMICIII embeddings (cc/mimic) (described in the next paragraph). By default, the code always initializes the word embedding of each unique word in the dictionary with a unique random vector. In alternative, we replace the random initialization with a pre-trained embedding. However, although such datasets generate good embeddings in many cases, for domain-specific tasks such as DNR and CCE they can suffer from some lack of vocabulary. As a matter of fact, in health corpora it is common to find very technical and unusual words which are specific to the health domain. If GloVe is trained only with general-purpose datasets, it is likely that such words will be missing and will still have to be assigned with random vectors. In all cases, the embeddings are updated during training by the backpropagation step.

In order to solve this problem, we have generated a new word embedding by re-training GloVe with a large health domain dataset called MIMIC-III \cite{johnson2016mimic}. This dataset contains records of 53,423 distinct hospital admissions of adults to an intensive care unit between 2001 and 2012. The data, structured in 26 tables, include information such as vital signs, observations of care providers, diagnostic codes etc. We expect such a dataset to contain many of the technical words from the health domain that may not appear in general-domain datasets, and as the size of MIMIC-III is sufficiently large, we should be able to extract meaningful vector representations for these words. As a first step, we have selected a subset of the tables and columns, and generated a new dataset where each selected cell together with the title of the corresponding column form a pseudo-sentence. As the next step, we have used this dataset to re-train GloVe, and concatenated these specialized word embeddings with the others to create vectors that contain information from both approaches. Obviously, there are words that appear in the general dataset, but not in MIMIC-III, and the vice versa. In such cases, the corresponding embedding is still assigned randomly. If a word does not appear in either dataset, we assign its whole embedding randomly. 

\subsection{Character-level embeddings}

Following Lample et al. \cite{lample2016neural} we also add character-level embeddings of the words. Such embeddings reflect the actual sequence of characters of a word and have proven to be useful for specific-domain tasks and morphologically-rich languages. Typically, they contribute to catching prefixes and suffixes which are frequent in the domain, and correctly classifying the corresponding words. As an example, a word ending in ``cycline" is very likely a drug name, and a character-level embedding could help classify it correctly even if the word was not present in the training vocabulary.  All the characters are initialized with a random embedding, and then the embeddings are passed character-by-character to a dedicated LSTM in both forward and backward order. The final outputs in the respective directions promise to be useful encodings of the ending and the beginning of the word. These character-level embeddings are integral part of the LSTM architecture and are not available in the CRF or other models. The character embeddings, too, are updated during training with backpropagation.

\subsection{Feature augmentation}

\begin{figure*}[t!]
	\centering\includegraphics[width=\textwidth]{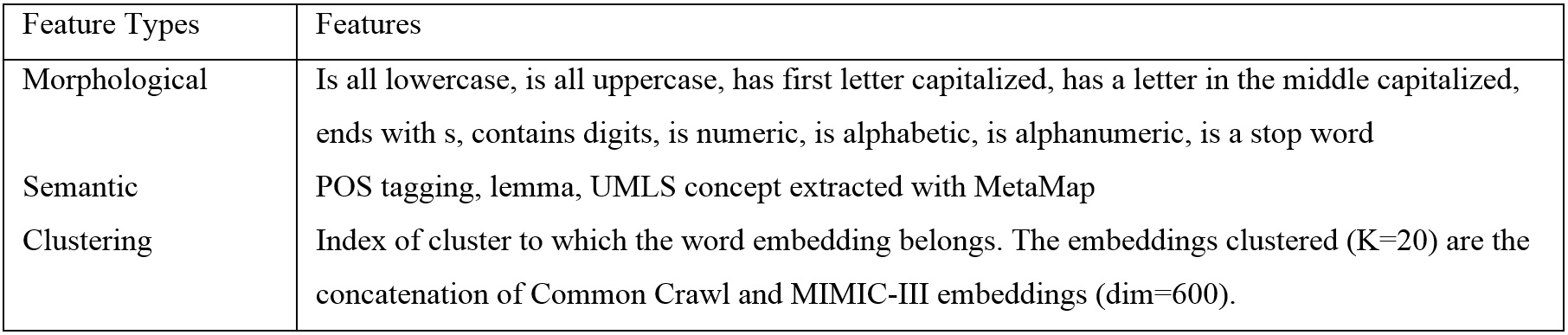}
	\caption{Description of the hand-crafted features.}
	\label{fig:fig2}
\end{figure*}

\begin{figure*}[t!]
	\centering\includegraphics[width=\textwidth]{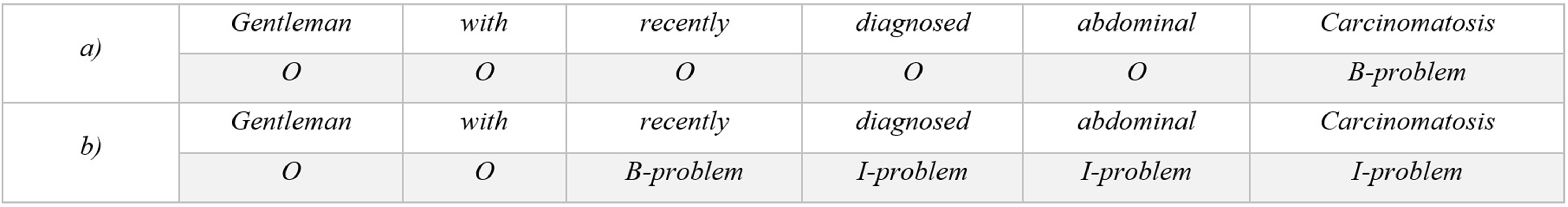}
	\caption{a) An example of an incorrect tagging in the ``strict" evaluation method. b) An example of a correct tagging in the ``strict" evaluation method}
	\label{fig:fig3}
\end{figure*}

Conventional machine learning approaches for NER usually have a feature engineering step. Lee et al. \cite{lee2016feature} have shown that adding hand-crafted features to a neural network can contribute to increase the recall. In our work, we try this approach with features similar to those used by Lee et al. \cite{lee2016feature}. Fig.\ref{fig:fig2} shows the list of features used. The distinct values of each feature are encoded onto short random vectors, for a total dimension of 146-D. During training, these encodings are updated as part of the backpropagation step.

\section{Results}
\label{S:5}

\subsection{Datasets}

Hereafter, we evaluate the models on three datasets in the health domain. The first is the \textit{2010 i2b2/VA IRB Revision} (we refer to it as \textit{i2b2/VA} for short in the following) and is used for evaluating CCE. This dataset is a reduced set of the original 2010 i2b2/VA dataset that is no longer distributed due to restrictions introduced by the Institutional Review Board (IRB) in 2011 \cite{uzuner20112010}. The other two datasets are \textit{DDI-DrugBank} and \textit{DDI-MedLine}, both part of the \textit{SemEval-2013 Task 9.1} for DNR \cite{herrero2013ddi}. Tables \ref{tab:1a} and \ref{tab:1b} describe the basic statistics of the datasets. For the experiments, we have used the official training and test splits released with the distributions.

\begin{table}[t]
\begin{subtable}[t]{0.5\textwidth}\centering
	{\begin{tabular}{ l | l l}
		\hline
		\textbf{} & Training & Test \\
		\hline
		Documents & 170 & 256 \\
		Sentences & 16315 & 27626 \\
		\hline
		\hline
		\textit{problem} & 7073 & 12592 \\
		\textit{test} & 4608 & 9225 \\
		\textit{treatment} & 4844 & 9344 \\
		\hline
	\end{tabular}}
	\caption{\textit{i2b2/VA}}\label{tab:1a}
	\vspace{5mm}
\end{subtable}

\begin{subtable}[t]{0.5\textwidth}\centering
	{\begin{tabular}{l | l l | l l}
			\hline
			\multirow{2}{*}{} &
				\multicolumn{2}{l|}{\textit{DrugBank}} &
				\multicolumn{2}{l}{\textit{MedLine}} \\
				& Training & Test & Training & Test \\
			\hline
			Documents & 730 & 54 & 175 & 58 \\
			Sentences & 6577 & 145 & 1627 & 520 \\
			\hline
			\hline
			\textit{drug\_n} & 124 & 6 & 520 & 115 \\
			\textit{group} & 3832 & 65 & 234 & 90 \\
			\textit{brand} & 1770 & 53 & 36 & 6 \\
			\textit{drug} & 9715 & 180 & 1574 & 171 \\
			\hline
	\end{tabular}}
	\caption{\textit{DrugBank} and \textit{MedLine}}\label{tab:1b}
\end{subtable}
\caption{Statistics of the training and test datasets used in the experiments}
\end{table}

\subsection{Evaluation metrics}

We report the performance of the model in terms of the F1-score. The F1-score is a very relevant measure as it considers both the precision and the recall, computing a weighted average of them. If we note as \textit{TP} the number of true positives, \textit{FP} the false positives and \textit{FN} the false negatives, we have:

\begin{equation}
\label{eq:f1_score}
\begin{aligned}
precision = \frac{TP}{TP+FP} \\
\vspace{5mm}
recall = \frac{TP}{TP+FN} \\
\vspace{2mm}
F1 = \frac{2*precision*recall}{precision+recall} \\
\end{aligned}
\end{equation}

However, it must be remarked that there are different ways of computing the precision and the recall, depending on what we consider as a correct or incorrect prediction \cite{nadeau2007survey}. In this work, we employ the ``strict'' evaluation method, where both the entity class and its exact boundaries are expected to be correct. We have used the B-I-O tagging standard to annotate the text at word level. In detail, `B' means the beginning (first word) of a named entity; `I' stands for `inside', meaning that the word is part of the same entity (for multi-word entities; e.g., ``albuterol sulfate''); and `O' stands for `outside', meaning that the word is not part of any named entities. Therefore, a valid annotation of a named entity always begins with a `B'. An example is shown in Fig.\ref{fig:fig3}. All the models used in this paper have been trained to predict explicit `B' and `I' labels for each entity class. The evaluation includes a preprocessing step that converts an `I' prediction to a `B' if it follows directly an `O' prediction, thus making all predicted entities valid. An entity is considered as correctly predicted only if all its `B' and `I' labels and all its classes are predicted correctly. In the example of Fig.\ref{fig:fig3} the prediction will be counted as a true positive only if all the four words ``recently diagnosed abdominal carcinomatosis'' are tagged as a single entity of the problem class. Every differing `B' prediction will instead be counted as a false positive. The evaluation protocol explicitly counts only the true positives and the false positives, and derives the false negatives as (number of true entities – true positives).


\subsection{Training and hyperparameters}

For an unbiased evaluation, all the trained models have been tested blindly on unseen test data. In order to facilitate replication of the empirical results, we have used a publicly-available library for the implementation of the neural networks (i.e. the Theano neural network toolkit \cite{bergstra2010theano}) and we release our code.\footnote{HealthNER: https://github.com/ijauregiCMCRC/healthNER} To operate, any machine learning model requires both a set of parameters, which are learned automatically during training, and some ``hyper-parameters", which have to be selected manually. Therefore, we have divided the training set of each dataset into two parts: a training set for learning the parameters (70\%), and a validation set (30\%) for selecting the best hyperparameters \cite{bergstra2012random}. The hyper-parameters of the LSTM include the number of hidden nodes (for both LSTM versions), $(H_w,H_c)\in\{25, 50, 100\}$; the word embedding dimension, $d_w \in \{50, 100, 200, 300, 600\}$; and the character embedding dimension, $d_c \in \{25, 50, 100\}$. Additional hyper-parameters include the learning rate and the drop-out rate, which were left to their default values of $[0.01]$ and $[0.5]$ respectively \cite{srivastava2014dropout}. All weights in the network, feature encodings and the words that do not have a pre-trained word embedding
have been initialized randomly from the uniform distribution within range [−1, 1], and updated during training with backpropagation. The number of training "epochs" (i.e., iterations) was set to 100, selecting the epoch that obtained the best results on the validation set. The best model from the validation set was finally tested on the unseen, independent test set without any further tuning, and the corresponding accuracy reported in the tables.  Table \ref{tab:hyper_params} shows all the hyper-parameters used for the experiments reported in the Results section.

\begin{table}[t]
	\centering
	\resizebox{0.45\textwidth}{!}{\begin{tabular}{l | l }
			\hline
			Hyper-parameter&Value\\
			\hline
			Word embedding dim ($d_w$)&300 (cc) or\\
			&600 (cc/mimic) \\
			\hline
			Word LSTM hidden layer dim ($H_w$)&100\\
			\hline
			Char embedding dim ($d_c$)&25\\
			\hline
			Char LSTM hidden layer dim ($H_c$)&25\\
			\hline
			Dropout&0.5\\
			\hline
			Optimization&Stochastic\\
			&Gradient Descend\\
			\hline
			Learning rate&0.01\\
			\hline
			Concatenated hand-crafted features dim&146\\
			\hline
	\end{tabular}}
	\caption{The hyper-parameters used in the experiments}\label{tab:hyper_params}
\end{table}

\begin{table}[t]
	\begin{subtable}[t]{0.5\textwidth}\centering
		\resizebox{0.80\textwidth}{!}{\begin{tabular}{ l | l }
				\hline
				Model & F1-score (\%) \\
				\hline
				Binarized Neural Embedding CRF \cite{wu2015study} & 82.80 \\
				CliNER \cite{boag2015cliner} & 80.00 \\
				Truecasing CRFSuite \cite{fu2014improving} & 75.86 \\
				\hline
				\hline
				CRF + (random) & 11.27 \\
				CRF + (features) & 25.53 \\
				CRF + (cc) & 53.72 \\
				CRF + (cc/mimic) & 58.28 \\
				CRF + (cc/mimic) + (features) & 64.09 \\
				\hline
				B-LSTM + (random) & 65.43 \\
				B-LSTM + (random) + (features) & 69.49\\
				B-LSTM + (cc) & 75.17 \\
				B-LSTM + (cc) + (char) & 76.79 \\
				B-LSTM + (cc/mimic) + (char) & 77.19 \\
				B-LSTM + (cc/mimic) + (char) + (features) & 77.59 \\
				\hline
				B-LSTM-CRF + (random) & 75.05 \\
				B-LSTM-CRF + (random) + (features) & 77.81 \\
				B-LSTM-CRF + (cc) & 82.85 \\
				\textbf{B-LSTM-CRF + (cc) + (char)} & \textbf{83.35} \\
				B-LSTM-CRF + (cc/mimic) + (char) & 82.70 \\
				B-LSTM-CRF + (cc/mimic) + (char) + (features) & 83.29 \\
				\hline		
		\end{tabular}}
		\caption{CCE results over the \textit{i2b2/VA} dataset}\label{tab:2a}
		\vspace{5mm}
	\end{subtable}
	
	\begin{subtable}[t]{0.5\textwidth}\centering
		\resizebox{0.80\textwidth}{!}{\begin{tabular}{l | l  | l }
				\hline
				\multirow{2}{*}{Model} &
				\textit{DrugBank} & \textit{MedLine} \\
				& F1(\%) & F1(\%) \\
				\hline
				WBI-NER \cite{rocktaschel2013wbi} & 87.80 & 58.10 \\
				Hybrid-DDI \cite{abacha2015text} & 80.00 & 37.00 \\
				Word2Vec+DINTO \cite{segura2015exploring} &	75.00 & 57.00 \\
				\hline
				\hline
				CRF + (random) & 28.70 & 13.65 \\
				CRF + (features) & 44.52 & 20.19 \\
				CRF + (cc) & 43.42 & 32.62 \\
				CRF + (cc/mimic) & 53.12 & 30.87 \\
				CRF + (cc/mimic) + (features) & 66.45 & 29.36 \\
				\hline
				B-LSTM + (random) & 65.09 & 21.28 \\
				B-LSTM + (random) + (features) & 75.43 & 30.88 \\
				B-LSTM+(cc) & 71.75 & 42.39 \\
				B-LSTM+(cc) + (char) & 84.35 & 43.33 \\
				B-LSTM + (cc/mimic) + (char) & 83.63 & 44.39 \\
				B-LSTM + (cc/mimic) + (char) + (features) & 84.06 & 45.92 \\
				\hline
				B-LSTM-CRF + (random) & 69.50 & 44.60 \\
				B-LSTM-CRF + (random) + (features) & 75.78 & 43.36 \\
				B-LSTM-CRF + (cc) & 79.03 & 57.87 \\
				B-LSTM-CRF + (cc) + (char) & 87.87 & 59.02 \\
				\textbf{B-LSTM-CRF + (cc/mimic) + (char)} & \textbf{88.38} & \textbf{60.66} \\
				B-LSTM-CRF + (cc/mimic) + (char) + (features) & 87.42 & 59.75 \\
				\hline
		\end{tabular}}
		\caption{DNR results over the \textit{DrugBank} and \textit{MedLine} datasets}\label{tab:2b}
	\end{subtable}
	\caption{Comparison of the results between the different RNN models and the state-of-the-art
		systems over the CNE and DNR tasks}
\end{table}

\subsection{Results}

Table \ref{tab:2a} and \ref{tab:2b} shows the results of the proposed models and the state-of-the-art systems on the CCE task (\textit{i2b2/VA} dataset) and DNR task (\textit{DrugBank} and \textit{MedLine} datasets), respectively. In the following subsections, we discuss the results obtained for each task.

\subsubsection{CCE results over the i2b2/VA dataset}

On the \textit{i2b2/VA} dataset (Table \ref{tab:2a}), the Bidirectional LSTM-CRF (B-LSTM-CRF) with Common Crawl embeddings (cc) and character-level embeddings (char) as features has obtained the best results (83.35\% F1-score). The model has outperformed all systems from the literature (top quadrant of Table \ref{tab:2a}) which are all based on conventional domain-specific feature engineering. It is important to note that deBruijin et al. \cite{de2011machine}  had reported a higher accuracy on 2010 i2b2/VA (85.23\% F1-score), but their model was trained and tested on the original version of the dataset which is no longer available due to the restrictions introduced by the Institutional Review Board. As for what specialized embeddings are concerned, Table \ref{tab:3} shows that the general-domain dataset Common Crawl already contains almost all the words in the dataset. Therefore, adding the MIMIC-III embeddings (mimic) does not extend the vocabulary, and it brings no improvement. On the other hand, the B-LSTM has improved by 0.3 pp with the cc/mimic embeddings. Even though the mimic embeddings do not cover significant extra vocabulary, they may have enriched the feature space. Conversely, the cc/mimic embeddings have provided no improvements
with the B-LSTM-CRF. For this, we need to take into account that the BLSTM-CRF already has a high score (83.35\% F1-score). Consequently, it may be more difficult to improve its results. Using conventional feature engineering has led to lower accuracy (77.81\% F1-score). Eventually, concatenating both the features and the pre-trained embeddings showed no improvement over the best model. Table \ref{tab:2a} also shows the importance of using a final CRF layer in the B-LSTM-CRF, given that the B-LSTM alone was only able to achieve a 77.59\% F1 score. At its turn, the CRF baseline has only obtained a 64.09\% F1 score in its best configuration, lower than any version of the LSTM.

\subsubsection{DNR results over the DrugBank and MedLine datasets}

In the DNR task (Table \ref{tab:2b}), the proposed B-LSTM-CRF with the concatenated word embeddings (cc/mimic) and the character-level embeddings (char) has improved over all the previous approaches on both \textit{DrugBank} (88.38\% F1 score) and \textit{MedLine} (60.66\% F1 score). Table \ref{tab:3} shows that only 49\% of the words in the datasets have been found in the cc embeddings. However, when the concatenated embeddings (cc/mimic) are used, the percentage of found words has increased to 67\% for \textit{DrugBank} and 61\% for \textit{MedLine}, leading to better results in the classification task. Words that appear in the MIMIC-III dataset but are not contained in Common Crawl are typically very technical and domain-specific, such as drug names or treatments; examples include: \textit{pentostatin}, \textit{sitagliptin}, \textit{hydrobromide}, \textit{organophosphate}, \textit{pyhisiological} and \textit{methimazole}. In total, 1189 extra words have been mapped in \textit{DrugBank} and 716 in \textit{MedLine} thanks to the use of MIMIC-III. However, the B-LSTM has only obtained an accuracy improvement on the \textit{MedLine} dataset, but not on \textit{DrugBank}. This can be explained by the fact that the accuracy of the B-LSTM on \textit{MedLine} is very low (44.33\%) and, therefore, easy to improve. Instead, on \textit{DrugBank} the accuracy of the B-LSTM is already very high (84.35\% F1-score) and thus difficult to improve. With the B-LSTM-CRF, results with extra vocabulary covered by the cc/mimic embeddings have improved with both datasets. 

As for what concerns the hand-crafted features, their use has led to higher accuracy than with the Common Crawl embeddings on the \textit{DrugBank} dataset in two cases. However, the concatenation of the features and the pre-trained embeddings has not improved the best results. As in the CCE task, the B-LSTM-CRF model has proved better than the B-LSTM alone on both \textit{DrugBank} (88.38\% vs 84.35\% F1-score) and \textit{MedLine} (60.66\% vs 45.92\% F1-score.) Finally, we can see that the use of the character-level embeddings has led to higher relative improvements for \textit{DrugBank} than for the other two datasets. A plausible explanation for this is that this dataset contains more words with distinctive prefixes and suffixes which are more effectively captured by the character-level embeddings. 

In general, the CRF has significantly underperformed compared to the neural networks. We speculate that this model may require more extensive feature engineering to achieve a comparable performance, or that it may not be able to achieve it at all. In particular, we see that the CRF has performed the worst with \textit{MedLine}. A possible explanation can be found in the ``curse of dimensionality'': \textit{MedLine} is a small dataset (1627 training sentences), while the overall dimensionality of the input embeddings is 746. This makes the learning problem very sparse and seems to seriously affect a linear model such as the CRF. On the contrary, the non-linear internal architecture of the neural networks may in some cases help reduce the effective dimensionality and mollify this problem.

\begin{table}[t]
	\centering
		\resizebox{0.45\textwidth}{!}{\begin{tabular}{l | l  | l  }
			\hline
			\multirow{3}{*}{} & 
			\multirow{2}{*}{\textit{Common Crawl}} & \textit{Common Crawl} \\
			& & \textit{+ MIMIC-III} \\
			& (cc) & (cc+mimic) \\
			\hline
			i2b2/VA & 99.99 \% & 99.99 \% \\
			DrugBank & 49.50 \% & 67.02 \% \\
			MedLine & 49.10 \% & 61.51 \% \\
			\hline
	\end{tabular}}
\caption{Percentage of words assigned with pre-trained embeddings in the train, dev and test of the respective datasets}\label{tab:3}
\end{table}

\begin{table*}[t]
	\begin{subtable}{\textwidth}\centering
		{\begin{tabular}{ l | l | l l l }
				\hline
				& Entity & Precision & Recall & F1-score \\
				\hline
				\multirow{3}{*}{B-LSTM-CRF+(cc)+(char)} &
				problem & 81.29 & 83.62 & 82.44 \\
				& test & 84.74 & 85.01 & 84.87 \\
				& treatment & 83.36 & 83.55 & 83.46 \\
				\hline	
		\end{tabular}}
		\caption{\textit{i2b2/VA}}\label{tab:4a}
		\vspace{5mm}
	\end{subtable}
	
	\begin{subtable}{\textwidth}\centering
		{\begin{tabular}{l | l | l  l  l | l  l  l }
				\hline
				\multirow{2}{*}{} &
				\multirow{2}{*}{Entity} & 
				\multicolumn{3}{l|}{\textit{DrugBank}} &
				\multicolumn{3}{l}{\textit{MedLine}} \\
				& & Precision & Recall & F1-score & Precision & Recall & F1-score \\
				\hline
				\multirow{2}{*}{B-LSTM-CRF} &
				group & 81.69 & 87.88 & 84.67 & 69.14 & 60.22 & 64.37 \\
				& drug & 94.77 & 89.56 & 91.83 & 73.89 & 77.33 & 75.57 \\
				+(cc+mimic) & brand & 84.21 & 90.57 & 87.27 & 100.00 & 16.67 & 28.57 \\
				+(char) & drug\_n & 00.00 & 00.00 & 00.00 & 68.18 & 25.57 & 37.19 \\
				\hline
		\end{tabular}}
		\caption{\textit{DrugBank} and \textit{MedLine}}\label{tab:4b}
	\end{subtable}
	\caption{Results by class for the B-LSTM-CRF with character-level and cc/mimic embeddings}
\end{table*}

\subsubsection{DNR results over the DrugBank and MedLine datasets}

Table \ref{tab:4a} and \ref{tab:4b} break down the results by entity class for the best model on each dataset. With the \textit{MedLine} dataset, we can notice the poor performance at detecting \textit{brand}. In \textit{DrugBank}, the same issue occurs with entity class \textit{drug\_n}. This issue is likely attributable to the small sample size. Instead, the \textit{i2b2/VA} dataset all entity classes are detected with similar F1-scores, likely owing to the larger number of samples per class. However, we see that \textit{brand} achieves the second best F1-score in \textit{DrugBank} despite its relatively low frequency in the dataset, and that \textit{drug\_n} obtains a very poor performance in \textit{MedLine} even if it has the second highest frequency. We identify two other main factors that may have a major impact on the accuracy: (1) the average length of the entities in each class, and (2) the number of test entities that had not been seen during the training stage. In this respect, the \textit{brand} and \textit{drug} entities are usually very short (average $\sim$ 1 word), while the \textit{group} and \textit{drug\_n} entities often have multiple words. Since shorter entities are easier to predict correctly, \textit{brand} obtains better accuracy than \textit{group} in \textit{DrugBank}. On the other hand, the \textit{drug\_n} and \textit{group} entities have similar length, but in \textit{MedLine} \textit{drug\_n} obtains a very poor performance. This is most likely because no entity of type \textit{drug\_n} that appears in the test set had been seen during training. Conversely, a large percentage of the test \textit{group} entities had been seen during training and have therefore proved easier to predict.

\section{Conclusion}
\label{S:6}

In this paper, we have set to investigate the effectiveness of the Bidirectional LSTM and Bidirectional LSTM-CRF --two specific architectures of recurrent neural networks-- for drug name recognition and clinical concept extraction, and compared them with a baseline CRF model. As input features, we have applied combinations of different word embeddings (Common Crawl and MIMIC-III), character-level embedding and conventional feature engineering. We have showed that the neural network models have obtained significantly better results than the CRF, and reported state-of-the-art results over the \textit{i2b2/VA}, \textit{DrugBank} and \textit{MedLine} datasets using the B-LSTM-CRF model. We have also provided evidence that retraining GloVe on a domain-specific dataset such as MIMIC-III can help learn vector representations for domain-specific words and increase the classification accuracy. Finally, we have showed that adding hand-crafted features does not further improve performance since the neural networks can learn useful word representations automatically from pre-trained word embeddings. Consequently, time-consuming, domain-specific feature engineering can be usefully avoided.





\bibliographystyle{model1-num-names}
\bibliography{biblio}

\begin{thebibliography}{36}
\expandafter\ifx\csname natexlab\endcsname\relax\def\natexlab#1{#1}\fi
\providecommand{\bibinfo}[2]{#2}
\ifx\xfnm\relax \def\xfnm[#1]{\unskip,\space#1}\fi
\bibitem[{Segura-Bedmar et~al.(2015)Segura-Bedmar, Su{\'a}rez-Paniagua, and
  Mart{\i}nez}]{segura2015exploring}
\bibinfo{author}{I.~Segura-Bedmar}, \bibinfo{author}{V.~Su{\'a}rez-Paniagua},
  \bibinfo{author}{P.~Mart{\i}nez},
\newblock \bibinfo{title}{Exploring word embedding for drug name recognition},
\newblock in: \bibinfo{booktitle}{\textit{6th International Workshop on Health
  Text Mining and Information Analysis (LOUHI)}, (2015).}
\bibitem[{Abacha et~al.(2015)Abacha, Chowdhury, Karanasiou, Mrabet, Lavelli,
  and Zweigenbaum}]{abacha2015text}
\bibinfo{author}{A.~B. Abacha}, \bibinfo{author}{M.~F.~M. Chowdhury},
  \bibinfo{author}{A.~Karanasiou}, \bibinfo{author}{Y.~Mrabet},
  \bibinfo{author}{A.~Lavelli}, \bibinfo{author}{P.~Zweigenbaum},
\newblock \bibinfo{title}{Text mining for pharmacovigilance: Using machine
  learning for drug name recognition and drug--drug interaction extraction and
  classification},
\newblock \bibinfo{journal}{\textit{Journal of biomedical informatics}}
  \bibinfo{volume}{58} (\bibinfo{year}{2015}) \bibinfo{pages}{122--132}.
\bibitem[{Rockt{\"a}schel et~al.(2013)Rockt{\"a}schel, Huber, Weidlich, and
  Leser}]{rocktaschel2013wbi}
\bibinfo{author}{T.~Rockt{\"a}schel}, \bibinfo{author}{T.~Huber},
  \bibinfo{author}{M.~Weidlich}, \bibinfo{author}{U.~Leser},
\newblock \bibinfo{title}{Wbi-ner: The impact of domain-specific features on
  the performance of identifying and classifying mentions of drugs},
\newblock in: \bibinfo{booktitle}{\textit{Second Joint Conference on Lexical
  and Computational Semantics (* SEM)}}, volume \bibinfo{volume}{2, (2013)},
  pp. \bibinfo{pages}{356--363}.
\bibitem[{de~Bruijn et~al.(2011)de~Bruijn, Cherry, Kiritchenko, Martin, and
  Zhu}]{de2011machine}
\bibinfo{author}{B.~de~Bruijn}, \bibinfo{author}{C.~Cherry},
  \bibinfo{author}{S.~Kiritchenko}, \bibinfo{author}{J.~Martin},
  \bibinfo{author}{X.~Zhu},
\newblock \bibinfo{title}{Machine-learned solutions for three stages of
  clinical information extraction: the state of the art at i2b2 2010},
\newblock \bibinfo{journal}{\textit{Journal of the American Medical Informatics
  Association}} \bibinfo{volume}{18} (\bibinfo{year}{2011})
  \bibinfo{pages}{557--562}.
\bibitem[{Lample et~al.(2016)Lample, Ballesteros, Subramanian, Kawakami, and
  Dyer}]{lample2016neural}
\bibinfo{author}{G.~Lample}, \bibinfo{author}{M.~Ballesteros},
  \bibinfo{author}{S.~Subramanian}, \bibinfo{author}{K.~Kawakami},
  \bibinfo{author}{C.~Dyer},
\newblock \bibinfo{title}{Neural architectures for named entity recognition},
\newblock \bibinfo{journal}{\textit{arXiv preprint} arXiv:1603.01360}
  (\bibinfo{year}{2016}).
\bibitem[{Hinton et~al.(2012)Hinton, Deng, Yu, Dahl, Mohamed, Jaitly, Senior,
  Vanhoucke, Nguyen, Sainath et~al.}]{hinton2012deep}
\bibinfo{author}{G.~Hinton}, \bibinfo{author}{L.~Deng},
  \bibinfo{author}{D.~Yu}, \bibinfo{author}{G.~E. Dahl}, \bibinfo{author}{A.-r.
  Mohamed}, \bibinfo{author}{N.~Jaitly}, \bibinfo{author}{A.~Senior},
  \bibinfo{author}{V.~Vanhoucke}, \bibinfo{author}{P.~Nguyen},
  \bibinfo{author}{T.~N. Sainath}, et~al.,
\newblock \bibinfo{title}{Deep neural networks for acoustic modeling in speech
  recognition: The shared views of four research groups},
\newblock \bibinfo{journal}{\textit{IEEE Signal Processing Magazine}}
  \bibinfo{volume}{29} (\bibinfo{year}{2012}) \bibinfo{pages}{82--97}.
\bibitem[{Krizhevsky et~al.(2012)Krizhevsky, Sutskever, and
  Hinton}]{krizhevsky2012imagenet}
\bibinfo{author}{A.~Krizhevsky}, \bibinfo{author}{I.~Sutskever},
  \bibinfo{author}{G.~E. Hinton},
\newblock \bibinfo{title}{Imagenet classification with deep convolutional
  neural networks},
\newblock in: \bibinfo{booktitle}{\textit{Advances in neural information
  processing systems (NIPS)}, (2012)}, pp. \bibinfo{pages}{1097--1105}.
\bibitem[{Pennington et~al.(2014)Pennington, Socher, and
  Manning}]{pennington2014glove}
\bibinfo{author}{J.~Pennington}, \bibinfo{author}{R.~Socher},
  \bibinfo{author}{C.~D. Manning},
\newblock \bibinfo{title}{Glove: Global vectors for word representation.},
\newblock in: \bibinfo{booktitle}{\textit{Empirical Methods on Natural Language
  Processing (EMNLP)}}, volume \bibinfo{volume}{14, (2014)}, pp.
  \bibinfo{pages}{1532--1543}.
\bibitem[{Mikolov et~al.(2013)Mikolov, Sutskever, Chen, Corrado, and
  Dean}]{mikolov2013distributed}
\bibinfo{author}{T.~Mikolov}, \bibinfo{author}{I.~Sutskever},
  \bibinfo{author}{K.~Chen}, \bibinfo{author}{G.~S. Corrado},
  \bibinfo{author}{J.~Dean},
\newblock \bibinfo{title}{Distributed representations of words and phrases and
  their compositionality},
\newblock in: \bibinfo{booktitle}{\textit{Advances in neural information
  processing systems (NIPS)}, (2013)}, pp. \bibinfo{pages}{3111--3119}.
\bibitem[{Johnson et~al.(2016)Johnson, Pollard, Shen, Lehman, Feng, Ghassemi,
  Moody, Szolovits, Celi, and Mark}]{johnson2016mimic}
\bibinfo{author}{A.~E. Johnson}, \bibinfo{author}{T.~J. Pollard},
  \bibinfo{author}{L.~Shen}, \bibinfo{author}{L.-W.~H. Lehman},
  \bibinfo{author}{M.~Feng}, \bibinfo{author}{M.~Ghassemi},
  \bibinfo{author}{B.~Moody}, \bibinfo{author}{P.~Szolovits},
  \bibinfo{author}{L.~A. Celi}, \bibinfo{author}{R.~G. Mark},
\newblock \bibinfo{title}{Mimic-iii, a freely accessible critical care
  database},
\newblock \bibinfo{journal}{\textit{Scientific data}} \bibinfo{volume}{3}
  (\bibinfo{year}{2016}).
\bibitem[{Chalapathy et~al.(2016{\natexlab{a}})Chalapathy, Borzeshi, and
  Piccardi}]{chalapathy2016investigation}
\bibinfo{author}{R.~Chalapathy}, \bibinfo{author}{E.~Z. Borzeshi},
  \bibinfo{author}{M.~Piccardi},
\newblock \bibinfo{title}{An investigation of recurrent neural architectures
  for drug name recognition},
\newblock in: \bibinfo{booktitle}{\textit{7th International Workshop on Health
  Text Mining and Information Analysis (LOUHI)}, (2016)}.
\bibitem[{Chalapathy et~al.(2016{\natexlab{b}})Chalapathy, Borzeshi, and
  Piccardi}]{chalapathy2016bidirectional}
\bibinfo{author}{R.~Chalapathy}, \bibinfo{author}{E.~Z. Borzeshi},
  \bibinfo{author}{M.~Piccardi},
\newblock \bibinfo{title}{Bidirectional lstm-crf for clinical concept
  extraction},
\newblock in: \bibinfo{booktitle}{\textit{Clinical Natural Language Processing
  Workshop (ClinicalNLP)}, (2016)}.
\bibitem[{Liu et~al.(2015)Liu, Tang, Chen, Wang, and Fan}]{liu2015feature}
\bibinfo{author}{S.~Liu}, \bibinfo{author}{B.~Tang}, \bibinfo{author}{Q.~Chen},
  \bibinfo{author}{X.~Wang}, \bibinfo{author}{X.~Fan},
\newblock \bibinfo{title}{Feature engineering for drug name recognition in
  biomedical texts: Feature conjunction and feature selection},
\newblock \bibinfo{journal}{\textit{Computational and mathematical methods in
  medicine}}  (\bibinfo{year}{2015}).
\bibitem[{Boag et~al.(2015)Boag, Wacome, Naumann, and
  Rumshisky}]{boag2015cliner}
\bibinfo{author}{W.~Boag}, \bibinfo{author}{K.~Wacome},
  \bibinfo{author}{T.~Naumann}, \bibinfo{author}{A.~Rumshisky},
\newblock \bibinfo{title}{Cliner: A lightweight tool for clinical named entity
  recognition},
\newblock \bibinfo{journal}{\textit{AMIA Joint Summits on Clinical Research
  Informatics (poster)}}  (\bibinfo{year}{2015}).
\bibitem[{Lebret and Collobert(2013)}]{lebret2013word}
\bibinfo{author}{R.~Lebret}, \bibinfo{author}{R.~Collobert},
\newblock \bibinfo{title}{Word emdeddings through hellinger pca},
\newblock in: \bibinfo{booktitle}{\textit{European Chapter of the Association
  for Computational Linguistics (EACL)}, (2013)}.
\bibitem[{Nikfarjam et~al.(2015)Nikfarjam, Sarker, O’Connor, Ginn, and
  Gonzalez}]{nikfarjam2015pharmacovigilance}
\bibinfo{author}{A.~Nikfarjam}, \bibinfo{author}{A.~Sarker},
  \bibinfo{author}{K.~O’Connor}, \bibinfo{author}{R.~Ginn},
  \bibinfo{author}{G.~Gonzalez},
\newblock \bibinfo{title}{Pharmacovigilance from social media: mining adverse
  drug reaction mentions using sequence labeling with word embedding cluster
  features},
\newblock \bibinfo{journal}{\textit{Journal of the American Medical Informatics
  Association}}  (\bibinfo{year}{2015}).
\bibitem[{Wu et~al.(2015)Wu, Xu, Jiang, Zhang, and Xu}]{wu2015study}
\bibinfo{author}{Y.~Wu}, \bibinfo{author}{J.~Xu}, \bibinfo{author}{M.~Jiang},
  \bibinfo{author}{Y.~Zhang}, \bibinfo{author}{H.~Xu},
\newblock \bibinfo{title}{A study of neural word embeddings for named entity
  recognition in clinical text},
\newblock in: \bibinfo{booktitle}{\textit{AMIA Annual Symposium Proceedings},
  (2015)}, p. \bibinfo{pages}{1326}.
\bibitem[{Dernoncourt et~al.(2017)Dernoncourt, Lee, Uzuner, and
  Szolovits}]{dernoncourt2017identification}
\bibinfo{author}{F.~Dernoncourt}, \bibinfo{author}{J.~Y. Lee},
  \bibinfo{author}{{\"O}.~Uzuner}, \bibinfo{author}{P.~Szolovits},
\newblock \bibinfo{title}{De-identification of patient notes with recurrent
  neural networks},
\newblock \bibinfo{journal}{\textit{Journal of the American Medical Informatics
  Association}} \bibinfo{volume}{24} (\bibinfo{year}{2017})
  \bibinfo{pages}{596--606}.
\bibitem[{Cocos et~al.(2017)Cocos, Fiks, and Masino}]{cocos2017deep}
\bibinfo{author}{A.~Cocos}, \bibinfo{author}{A.~G. Fiks},
  \bibinfo{author}{A.~J. Masino},
\newblock \bibinfo{title}{Deep learning for pharmacovigilance: recurrent neural
  network architectures for labeling adverse drug reactions in twitter posts},
\newblock \bibinfo{journal}{Journal of the American Medical Informatics
  Association} \bibinfo{volume}{24} (\bibinfo{year}{2017})
  \bibinfo{pages}{813--821}.
\bibitem[{Xie et~al.(2017)Xie, Liu, and Dajun~Zeng}]{xie2017mining}
\bibinfo{author}{J.~Xie}, \bibinfo{author}{X.~Liu},
  \bibinfo{author}{D.~Dajun~Zeng},
\newblock \bibinfo{title}{Mining e-cigarette adverse events in social media
  using bi-lstm recurrent neural network with word embedding representation},
\newblock \bibinfo{journal}{Journal of the American Medical Informatics
  Association} \bibinfo{volume}{25} (\bibinfo{year}{2017})
  \bibinfo{pages}{72--80}.
\bibitem[{Wei et~al.(2016)Wei, Chen, Xu, He, and Gui}]{wei2016disease}
\bibinfo{author}{Q.~Wei}, \bibinfo{author}{T.~Chen}, \bibinfo{author}{R.~Xu},
  \bibinfo{author}{Y.~He}, \bibinfo{author}{L.~Gui},
\newblock \bibinfo{title}{Disease named entity recognition by combining
  conditional random fields and bidirectional recurrent neural networks},
\newblock \bibinfo{journal}{Database}  (\bibinfo{year}{2016}).
\bibitem[{Jagannatha and Yu(2016)}]{jagannatha2016structured}
\bibinfo{author}{A.~N. Jagannatha}, \bibinfo{author}{H.~Yu},
\newblock \bibinfo{title}{Structured prediction models for rnn based sequence
  labeling in clinical text},
\newblock in: \bibinfo{booktitle}{Proceedings of the Conference on Empirical
  Methods in Natural Language Processing. Conference on Empirical Methods in
  Natural Language Processing}, volume \bibinfo{volume}{2016},
  \bibinfo{organization}{NIH Public Access}, p. \bibinfo{pages}{856}.
\bibitem[{Gridach(2017)}]{gridach2017character}
\bibinfo{author}{M.~Gridach},
\newblock \bibinfo{title}{Character-level neural network for biomedical named
  entity recognition},
\newblock \bibinfo{journal}{Journal of biomedical informatics}
  \bibinfo{volume}{70} (\bibinfo{year}{2017}) \bibinfo{pages}{85--91}.
\bibitem[{Lee et~al.(2016)Lee, Dernoncourt, Uzuner, and
  Szolovits}]{lee2016feature}
\bibinfo{author}{J.~Y. Lee}, \bibinfo{author}{F.~Dernoncourt},
  \bibinfo{author}{O.~Uzuner}, \bibinfo{author}{P.~Szolovits},
\newblock \bibinfo{title}{Feature-augmented neural networks for patient note
  de-identification},
\newblock in: \bibinfo{booktitle}{\textit{Clinical Natural Language Processing
  Workshop (ClinicalNLP)}, (2016)}.
\bibitem[{Lafferty et~al.(2001)Lafferty, McCallum, Pereira
  et~al.}]{lafferty2001conditional}
\bibinfo{author}{J.~Lafferty}, \bibinfo{author}{A.~McCallum},
  \bibinfo{author}{F.~Pereira}, et~al.,
\newblock \bibinfo{title}{Conditional random fields: Probabilistic models for
  segmenting and labeling sequence data},
\newblock in: \bibinfo{booktitle}{\textit{Proceedings of the eighteenth
  international conference on machine learning, ICML}, (2001)},
  volume~\bibinfo{volume}{1}, pp. \bibinfo{pages}{282--289}.
\bibitem[{HCR(2017)}]{HCRF}
\bibinfo{title}{Hcrf},
  \bibinfo{howpublished}{\url{http://multicomp.ict.usc.edu/?p=790}},
  \bibinfo{year}{2017}. \bibinfo{note}{Accessed: 2017-05-01}.
\bibitem[{Bengio et~al.(1994)Bengio, Simard, and Frasconi}]{bengio1994learning}
\bibinfo{author}{Y.~Bengio}, \bibinfo{author}{P.~Simard},
  \bibinfo{author}{P.~Frasconi},
\newblock \bibinfo{title}{Learning long-term dependencies with gradient descent
  is difficult},
\newblock \bibinfo{journal}{\textit{IEEE transactions on neural networks}}
  \bibinfo{volume}{5} (\bibinfo{year}{1994}) \bibinfo{pages}{157--166}.
\bibitem[{Hochreiter and Schmidhuber(1997)}]{hochreiter1997long}
\bibinfo{author}{S.~Hochreiter}, \bibinfo{author}{J.~Schmidhuber},
\newblock \bibinfo{title}{Long short-term memory},
\newblock \bibinfo{journal}{\textit{Neural computation}} \bibinfo{volume}{9}
  (\bibinfo{year}{1997}) \bibinfo{pages}{1735--1780}.
\bibitem[{Glo(2014)}]{GloVe}
\bibinfo{title}{Glove},
  \bibinfo{howpublished}{\url{https://nlp.stanford.edu/projects/glove/}},
  \bibinfo{year}{2014}. \bibinfo{note}{Accessed: 2017-05-01}.
\bibitem[{Uzuner et~al.(2011)Uzuner, South, Shen, and DuVall}]{uzuner20112010}
\bibinfo{author}{{\"O}.~Uzuner}, \bibinfo{author}{B.~R. South},
  \bibinfo{author}{S.~Shen}, \bibinfo{author}{S.~L. DuVall},
\newblock \bibinfo{title}{2010 i2b2/va challenge on concepts, assertions, and
  relations in clinical text},
\newblock \bibinfo{journal}{\textit{Journal of the American Medical Informatics
  Association}} \bibinfo{volume}{18} (\bibinfo{year}{2011})
  \bibinfo{pages}{552--556}.
\bibitem[{Herrero-Zazo et~al.(2013)Herrero-Zazo, Segura-Bedmar, Mart{\'\i}nez,
  and Declerck}]{herrero2013ddi}
\bibinfo{author}{M.~Herrero-Zazo}, \bibinfo{author}{I.~Segura-Bedmar},
  \bibinfo{author}{P.~Mart{\'\i}nez}, \bibinfo{author}{T.~Declerck},
\newblock \bibinfo{title}{The ddi corpus: An annotated corpus with
  pharmacological substances and drug--drug interactions},
\newblock \bibinfo{journal}{\textit{Journal of biomedical informatics}}
  \bibinfo{volume}{46} (\bibinfo{year}{2013}) \bibinfo{pages}{914--920}.
\bibitem[{Nadeau and Sekine(2007)}]{nadeau2007survey}
\bibinfo{author}{D.~Nadeau}, \bibinfo{author}{S.~Sekine},
\newblock \bibinfo{title}{A survey of named entity recognition and
  classification},
\newblock \bibinfo{journal}{\textit{Lingvisticae Investigationes}}
  \bibinfo{volume}{30} (\bibinfo{year}{2007}) \bibinfo{pages}{3--26}.
\bibitem[{Bergstra et~al.(2010)Bergstra, Breuleux, Bastien, Lamblin, Pascanu,
  Desjardins, Turian, Warde-Farley, and Bengio}]{bergstra2010theano}
\bibinfo{author}{J.~Bergstra}, \bibinfo{author}{O.~Breuleux},
  \bibinfo{author}{F.~Bastien}, \bibinfo{author}{P.~Lamblin},
  \bibinfo{author}{R.~Pascanu}, \bibinfo{author}{G.~Desjardins},
  \bibinfo{author}{J.~Turian}, \bibinfo{author}{D.~Warde-Farley},
  \bibinfo{author}{Y.~Bengio},
\newblock \bibinfo{title}{Theano: A cpu and gpu math compiler in python},
\newblock in: \bibinfo{booktitle}{\textit{Proceedings of the 9th Python in
  Science Conference}, (2010)}, pp. \bibinfo{pages}{1--7}.
\bibitem[{Bergstra and Bengio(2012)}]{bergstra2012random}
\bibinfo{author}{J.~Bergstra}, \bibinfo{author}{Y.~Bengio},
\newblock \bibinfo{title}{Random search for hyper-parameter optimization},
\newblock \bibinfo{journal}{\textit{Journal of Machine Learning Research}}
  \bibinfo{volume}{13} (\bibinfo{year}{2012}) \bibinfo{pages}{281--305}.
\bibitem[{Srivastava et~al.(2014)Srivastava, Hinton, Krizhevsky, Sutskever, and
  Salakhutdinov}]{srivastava2014dropout}
\bibinfo{author}{N.~Srivastava}, \bibinfo{author}{G.~Hinton},
  \bibinfo{author}{A.~Krizhevsky}, \bibinfo{author}{I.~Sutskever},
  \bibinfo{author}{R.~Salakhutdinov},
\newblock \bibinfo{title}{Dropout: A simple way to prevent neural networks from
  overfitting},
\newblock \bibinfo{journal}{\textit{The Journal of Machine Learning Research}}
  \bibinfo{volume}{15} (\bibinfo{year}{2014}) \bibinfo{pages}{1929--1958}.
\bibitem[{Fu and Ananiadou(2014)}]{fu2014improving}
\bibinfo{author}{X.~Fu}, \bibinfo{author}{S.~Ananiadou},
\newblock \bibinfo{title}{Improving the extraction of clinical concepts from
  clinical records},
\newblock \bibinfo{journal}{\textit{Proceedings of BioTxtM14}}
  (\bibinfo{year}{2014}).

\end{thebibliography}






\end{document}